\documentclass[10pt,twocolumn,letterpaper]{article}
\usepackage{cvpr}

\def\paperID{****}
\def\confName{MMFM-BIOMED @ CVPR}
\def\confYear{2026}

\newcommand{\para}[1]{\vspace{3pt}\noindent\textbf{#1.}\;}
\usepackage{microtype}
\usepackage[letterpaper,top=1in,bottom=1.125in,left=0.8125in,right=0.8125in,columnsep=0.375in]{geometry}
\usepackage{hyperref}
\hypersetup{colorlinks=true,linkcolor=black,citecolor=black,urlcolor=blue}

\begin{document}
\hyphenation{slice-based multi-site re-spec-ialised trans-form-ers
             tem-po-ro-man-dib-u-lar os-teo-ar-thri-tis
             back-bone back-bones}

\sloppy

\title{%
  \vspace{-6pt}
  Self-Supervised Vision Transformers for CBCT-Based\\
  Detection of Temporomandibular Joint Osteoarthritis
  \vspace{-4pt}
}
\author{
  Shradhdha Trivedi$^{*}$ \quad Vrundan Sojitra$^{\dagger}$ \quad Mariela Padilla$^{*}$
  \\[4pt]
  {\small $^{*}$Herman Ostrow School of Dentistry, University of Southern California}
  \\
  {\small $^{\dagger}$Viterbi School of Engineering, University of Southern California}
  \vspace{-6pt}
}
\maketitle
\thispagestyle{empty}

\begin{abstract}
Temporomandibular joint osteoarthritis (TMJ OA) is a prevalent
degenerative condition whose osseous changes are often subtle on
cone-beam CT (CBCT), making automated detection challenging.
We study how well the DINO family of self-supervised vision
transformers---DINOv1, DINOv2, DINOv2+reg, and RAD-DINO (a
radiology-pretrained variant)---transfers to CBCT, asking how much
backbone adaptation is needed and of what kind.
We propose a simple slice-based pipeline using Vision Transformer
(ViT) backbones: axial CBCT slices are encoded per-slice by a frozen
or partially adapted ViT and aggregated via attention-based multiple
instance learning (MIL) for patient-level binary OA\,/\,Normal
classification.
Through systematic ablation across unfreezing strategies and
aggregation designs on a multi-source CBCT dataset, we find that
partial unfreezing of the final two transformer blocks is the
decisive factor, improving AUC from 0.671 (fully frozen DINOv2)
to 0.902. This outperforms DINOv1 (0.867), DINOv2+reg (0.774),
and a supervised ImageNet ViT-B/16 baseline (0.843).
Our results provide practical guidance for adapting DINO-family
foundation models in low-data medical imaging settings, showing
that adaptation strategy is a stronger driver of performance than
backbone choice alone.
\end{abstract}

\section{Introduction}
\label{sec:intro}

Temporomandibular joint osteoarthritis (TMJ OA) affects a significant
proportion of adults and is characterised by progressive osseous
remodelling of the mandibular condyle—erosion, flattening,
osteophyte formation, and subchondral sclerosis~\cite{schiffman2014}.
Cone-beam CT (CBCT) is the primary modality for evaluating osseous
TMJ morphology in dental practice due to its high spatial resolution
and accessibility. Yet radiographic features of early-to-moderate OA
are subtle, leading to inter-observer variability and delayed
treatment~\cite{xu2023}.

Automated deep learning classifiers for TMJ OA from CBCT have been
proposed~\cite{lee2020,jung2023,eser2023}, but rely on fully
supervised training with manually cropped ROIs and show limited
generalisation across sites. Large-scale self-supervised vision
foundation models—particularly the DINO family~\cite{caron2021dino,
oquab2024dinov2}—learn rich transferable visual representations
without labels, making them attractive for low-data medical imaging
settings where annotation is expensive.

Despite this promise, \emph{no prior work} systematically evaluates
how DINO-family ViTs should be adapted for CBCT—a domain that differs
fundamentally from natural images in contrast mechanism, noise
characteristics, and anatomical structure. Naive transfer (\ie
frozen backbone) may fail, while aggressive fine-tuning risks
overfitting in small datasets. The optimal adaptation regime is
unknown.

\para{Contributions}
\begin{itemize}
  \item To our knowledge, we provide the first systematic comparison of DINOv1,
        DINOv2, and DINOv2+reg adaptation strategies for CBCT-based
        binary TMJ OA detection, using a controlled multi-source
        evaluation protocol.
  \item We demonstrate that \emph{partial} backbone adaptation
        (last two transformer blocks) is the critical design choice
        across all backbones, consistently outperforming frozen
        representations and a supervised ImageNet ViT-B/16.
  \item We benchmark against a supervised ImageNet-pretrained ViT-B/16~\cite{dosovitskiy2021vit} under both frozen and partial adaptation,
        establishing a meaningful comparison absent from prior TMJ OA
        work.
  \item We provide an ablation of aggregation strategy, slice
        orientation, and slice count, giving practical guidance for
        CBCT foundation model deployment.
\end{itemize}

\section{Related Work}
\label{sec:related}

\para{TMJ OA detection from CBCT}
Supervised CNNs on 2D sagittal CBCT slices achieve strong performance
for radiographically apparent OA~\cite{lee2020}, but sensitivity drops
for borderline cases~\cite{jung2023}. Feature-pyramid approaches
(\eg YOLOv5~\cite{eser2023}) show high class-wise F1 but assume
pre-selected ROIs. All prior work is fully supervised and single-site.

\para{DINO-family self-supervised ViTs}
DINO~\cite{caron2021dino} introduces self-distillation with
momentum-updated teacher ViTs, yielding emergent segmentation
properties in the \texttt{[CLS]} attention maps.
DINOv2~\cite{oquab2024dinov2} scales training to LVD-142M images with
an iBOT masked-image-modelling auxiliary loss and SwAV-style
regularisation, substantially improving dense prediction transfer.
A register-token variant~\cite{darcet2024registers} (which we call DINOv2+reg) reduces
artefact attention on low-information patches, potentially beneficial
for the homogeneous background of CBCT scans.
Wolf~\etal~\cite{wolf2023} show SSL pretraining is most beneficial
when fewer than 200 annotated cases are available—directly relevant
to our setting.

\para{MIL for radiology}
Attention-based MIL~\cite{ilse2018} enables bag-level classification
without patch labels. Its combination with pretrained ViT features has
shown strong results in computational
pathology~\cite{chen2022hierarchical}; our work extends this paradigm
to 3D-to-2D CBCT analysis.

\section{Method}
\label{sec:method}

\begin{figure*}[t]
  \centering
  \includegraphics[width=\textwidth]{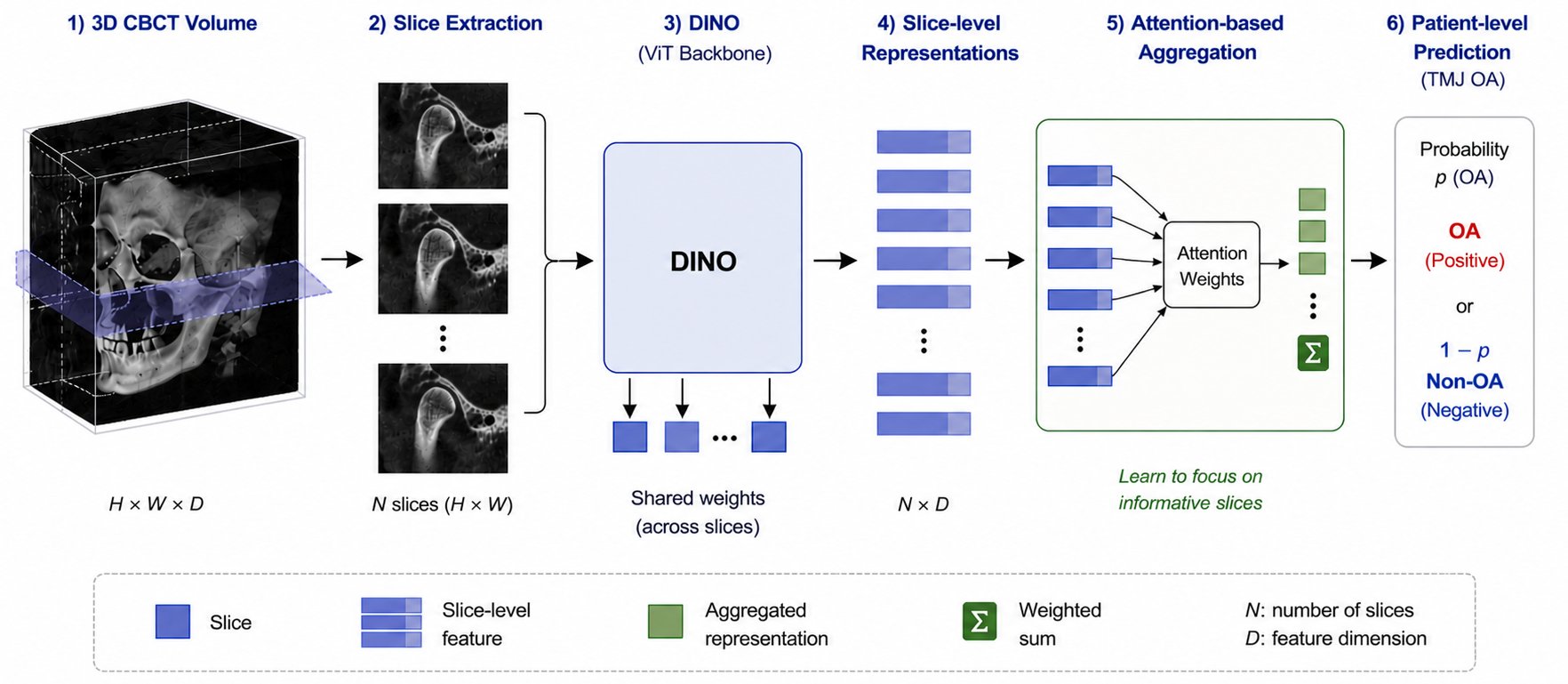}
  \vspace{-10pt}
  \caption{%
    \textbf{Proposed pipeline.}
    (1)~A 3D CBCT volume ($H\!\times\!W\!\times\!D$) is input.
    (2)~$N$ axial slices are extracted from the TMJ region.
    (3)~Each slice is independently encoded by a shared pretrained
        DINO-family ViT backbone (weights partially adapted).
    (4)~Per-slice \texttt{[CLS]} embeddings ($N\!\times\!D$) are
        formed.
    (5)~Attention-based MIL aggregation learns to up-weight
        diagnostically informative slices.
    (6)~A linear classifier outputs binary TMJ OA probability.
  }
  \label{fig:pipeline}
\end{figure*}

Figure~\ref{fig:pipeline} summarises the pipeline.

\subsection{CBCT Preprocessing and Slice Extraction}

Multi-site CBCT volumes are resampled to a consistent isotropic voxel
spacing and intensity-normalised per volume. We extract $N$ evenly
spaced 2D slices along the axial orientation, centred on the TMJ
region of each volume, assuming consistent anatomical positioning
of the TMJ across CBCT acquisitions.
Slices are resized to $518\!\times\!518$ for DINOv2 and
$224\!\times\!224$ for DINOv1, DINOv2+reg, and the ImageNet ViT baseline,
and channel-triplicated.
We ablate slice count ($N \in \{12, 24\}$) and orientation (axial vs.\
sagittal).

\subsection{DINO-Family Feature Extraction}

Each 2D slice is independently encoded by a ViT-B backbone to
produce a \texttt{[CLS]} token $\mathbf{f}_i \in \mathbb{R}^{768}$.
We evaluate three backbone variants:

\begin{itemize}
  \item \textbf{DINOv1}~\cite{caron2021dino}: ViT-B/16, self-supervised
        via self-distillation on ImageNet-1k.
  \item \textbf{DINOv2}~\cite{oquab2024dinov2}: ViT-B/14, trained on
        LVD-142M with iBOT and SwAV objectives.
  \item \textbf{DINOv2+reg}~\cite{darcet2024registers}: a variant
        incorporating register tokens, which suppress artefact
        attention on uninformative background regions. We refer
        to this as ``DINOv3-style'' for clarity throughout.
  \item \textbf{RAD-DINO}~\cite{perez2024raddino}: a DINOv2-based
        model further pretrained on large-scale radiology images
        (chest X-rays and CT), providing domain-specific
        initialisation closer to medical imaging.
\end{itemize}

We also include a \textbf{supervised ImageNet ViT-B/16} as a
non-SSL baseline.

\para{Adaptation strategies (ablation)}
\begin{itemize}
  \item \textbf{Frozen}: all backbone weights fixed; only the MIL
        head is trained.
  \item \textbf{Partial} (proposed): the last two transformer blocks
        and final LayerNorm are unfrozen with a reduced backbone
        LR of $5\!\times\!10^{-5}$ (10$\times$ smaller than the head).
\end{itemize}

\subsection{Attention-Based MIL Aggregation}

Patient-level prediction uses attention-pooling over the $N$
per-slice embeddings:
\begin{equation}
  a_i = \operatorname{softmax}_i\!\bigl(\mathbf{w}^\top
        \tanh(\mathbf{V}\mathbf{f}_i)\bigr),
  \qquad
  \mathbf{z} = \textstyle\sum_i a_i\,\mathbf{f}_i,
  \label{eq:mil}
\end{equation}
with $\mathbf{V}\!\in\!\mathbb{R}^{128\times768}$,
$\mathbf{w}\!\in\!\mathbb{R}^{128}$ learned.
Patient embedding $\mathbf{z}$ is passed to a linear classifier with
sigmoid output (binary OA\,/\,Normal).
We ablate mean pooling as a simpler alternative.

\para{Training}
All models are trained with AdamW, a ReduceLROnPlateau scheduler
(patience 5--7, factor 0.4--0.5), and early stopping (patience 10--15
epochs), for up to 30 epochs. Batch size is 8--16 patients.
Binary cross-entropy with positive-class weighting addresses label
imbalance; label smoothing (0--0.1) is applied where noted.
Dropout (0.3) is applied before the classifier head.
Augmentation: random horizontal flip and $\pm10^{\circ}$ rotation.

\section{Experiments}
\label{sec:experiments}

\subsection{Dataset}

This retrospective study was approved by the Institutional Review
Board (IRB) at the University of Southern California
(IRB \#HS-25-00599). Informed consent was waived given the
retrospective nature of the study and the use of de-identified
imaging data.

We combine an internal institutional CBCT cohort with the public
University of Michigan Deep Blue TMJ OA repository~\cite{cevidanes2024},
totalling 210 patients. Left and right TMJs are treated as
independent condyle-level samples. Splits are performed at the
\emph{patient} level, ensuring both condyles from the same patient
always reside in the same split and preventing data leakage
(Table~\ref{tab:dataset}). Labels are binary: OA-positive
(\emph{any} grade) vs.\ Normal.

\begin{table}[h]
  \vspace{-4pt}
  \centering
  \small
  \setlength{\tabcolsep}{5pt}
  \caption{Dataset splits (condyles\,/\,slices at $N{=}24$).}
  \label{tab:dataset}
  \begin{tabular}{lccc}
    \toprule
    Site & Train & Validation & Test \\
    \midrule
    Internal  & 170\,/\,4080 & 60\,/\,1440 & 50\,/\,1200 \\
    Deep Blue & 80\,/\,1920  & 20\,/\,480  & 40\,/\,960  \\
    \midrule
    Total     & 250\,/\,6000 & 80\,/\,1920 & 90\,/\,2160 \\
    \bottomrule
  \end{tabular}
  \vspace{-6pt}
\end{table}

\subsection{Main Ablation: Backbone and Adaptation}

Table~\ref{tab:main} reports AUC and F1 on the held-out validation set.

\begin{table}[t]
  \centering
  \small
  \setlength{\tabcolsep}{5pt}
  \caption{%
    \textbf{Main ablation: backbone and adaptation strategy.}
    All use attention-MIL, axial slices ($N{=}24$).
    DINOv2 is evaluated at both native (518px) and standard (224px)
    resolution. Best in \textbf{bold}.
  }
  \label{tab:main}
  \begin{tabular}{lcc}
    \toprule
    Model & AUC & F1 \\
    \midrule
    ImageNet ViT-B/16 — frozen  & 0.657 & 0.560 \\
    ImageNet ViT-B/16 — partial & 0.843 & 0.670 \\
    \midrule
    DINOv1 — frozen             & 0.714 & 0.667 \\
    DINOv1 — partial            & 0.867 & 0.800 \\
    \midrule
    DINOv2+reg — frozen         & 0.621 & 0.615 \\
    DINOv2+reg — partial        & 0.774 & 0.683 \\
    \midrule
    DINOv2 (518px) — frozen     & 0.671 & 0.714 \\
    DINOv2 (518px) — partial    & \textbf{0.902} & \textbf{0.889} \\
    DINOv2 (224px) — frozen     & 0.781 & 0.692 \\
    DINOv2 (224px) — partial    & 0.900 & 0.741 \\
    \midrule
    RAD-DINO — frozen           & 0.738 & 0.421 \\
    \bottomrule
  \end{tabular}
\end{table}

\para{Key findings}
\textbf{(1) Partial adaptation consistently improves all backbones.}
Unfreezing the last two transformer blocks improves AUC across every
model: DINOv1 (0.714\,$\to$\,0.867), DINOv2 (0.671\,$\to$\,0.902),
and DINOv2+reg (0.621\,$\to$\,0.774). Fully frozen representations are
insufficient for CBCT regardless of pretraining quality.

\textbf{(2) DINOv2 leads all backbones.}
DINOv2 with partial adaptation (AUC\,=\,0.902) outperforms DINOv1
(0.867), DINOv2+reg (0.774), and the supervised ImageNet ViT-B/16
(partial: 0.843, frozen: 0.657). Partial adaptation yields consistent
gains across all models, with the largest absolute improvement seen
for the supervised ImageNet baseline (+0.186), confirming that
SSL pretraining provides a better starting point for CBCT transfer.

\textbf{(3) Attention MIL and axial orientation are complementary.}
Mean pooling drops AUC by 5.4 points (0.902\,$\to$\,0.848) relative
to attention MIL on the same backbone, and sagittal slices (0.833)
underperform axial (0.902). Reducing slices from 24 to 12 incurs
only a modest drop (0.886), suggesting dense sampling is not critical.

\textbf{(4) Domain-specific pretraining helps, but adaptation matters more.}
RAD-DINO (frozen AUC 0.738) outperforms all frozen general-purpose
models without any task-specific adaptation, confirming that
radiology pretraining yields more transferable CBCT features.
However, partially adapted DINOv2 (0.902) surpasses frozen RAD-DINO
by 0.164 AUC points, demonstrating that \emph{how} a model is
adapted matters more than pretraining domain alone.

\textbf{(5) Adaptation strategy dominates over input resolution.}
To isolate the effect of resolution, we evaluate DINOv2 at both
518\,px and 224\,px. Frozen performance is higher at 224\,px
(AUC 0.781 vs.\ 0.671), suggesting that 224\,px inputs better align
with the pretraining scale of the ViT patch embedding.
Critically, partially fine-tuned models achieve comparable
performance at both resolutions (AUC 0.902 vs.\ 0.900), confirming
that \emph{adaptation strategy, not input resolution, is the dominant
factor}. This controlled experiment rules out resolution as a
confounding variable in our backbone comparisons.

\subsection{Aggregation and Slice Design}

\begin{table}[h]
  \vspace{-4pt}
  \centering
  \small
  \setlength{\tabcolsep}{5pt}
  \caption{Aggregation and slice ablation (DINOv2, partial adapt).}
  \label{tab:agg}
  \begin{tabular}{llcc}
    \toprule
    Aggregation & Orientation ($N$) & AUC & F1 \\
    \midrule
    Mean pool     & Axial (24)    & 0.848 & 0.857 \\
    Attention MIL & Sagittal (24) & 0.833 & 0.813 \\
    Attention MIL & Axial (12)    & 0.886 & 0.839 \\
    Attention MIL & Axial (24)    & \textbf{0.902} & \textbf{0.889} \\
    \bottomrule
  \end{tabular}
  \vspace{-4pt}
\end{table}

Attention MIL consistently outperforms mean pooling (AUC 0.902
vs.\ 0.848), suggesting that not all slices contribute equally to
the classification decision.
Axial slices outperform sagittal (0.902 vs.\ 0.833), and reducing
slice count from 24 to 12 incurs only a modest AUC drop (0.886),
indicating that dense volumetric sampling is not necessary.

\section{Discussion}
\label{sec:discussion}

\para{Why partial adaptation is the sweet spot}
Early ViT blocks encode low-level texture and structural features
(\eg bone edge contrast, trabecular patterns) that transfer
well from natural images to CBCT.
Final blocks encode task-spe\-cific semantic representations that
must be re-spec\-ialised to condylar morphology.
Adapting only blocks 11–12 achieves this re-specialisation while
preserving the stable feature basis of earlier layers—avoiding the
catastrophic forgetting that can arise when adapting large
medical datasets.

\para{RAD-DINO and domain-specific pretraining}
RAD-DINO's stronger frozen performance (0.738) relative to frozen
general-purpose models suggests radiology pretraining captures
bone texture and anatomical features more transferable to CBCT.
Yet the 0.164 AUC gap versus partially adapted DINOv2 shows that
targeted fine-tuning provides complementary benefits that
domain-specific pretraining alone cannot replicate. Partial
adaptation of RAD-DINO is a promising direction for future work.

\para{Threshold selection matters clinically}
At the default 0.5 threshold, DINOv2 partial shows high specificity
(0.933) but lower sensitivity (0.429), reflecting class imbalance
in the dataset.
At the optimal decision threshold (0.292), F1 rises to 0.889.
This operating point shift substantially improves sensitivity
without a disproportionate loss of specificity, underscoring
the importance of threshold tuning for clinical deployment.
In a screening setting, where missed OA cases carry higher cost
than false positives, operating below 0.5 is likely preferable
and should be determined in consultation with clinical partners.

\para{Limitations}
The dataset size (210 patients across two sites) limits statistical
power and may not reflect the full variability of TMJ morphology
across populations and scanners.
The 2D slice-based approach discards inter-slice spatial context
that may carry diagnostic information. Additionally, while DINOv2 was evaluated at both 518\,px and
224\,px resolutions (confirming resolution is not the key driver),
DINOv1, DINOv2+reg, and the ImageNet baseline were only evaluated at
224\,px; evaluating these at their native resolutions remains future
work. External multi-institutional validation is also an important
next step.

\section{Conclusion}
\label{sec:conclusion}

We present a systematic evaluation of pretrained vision transformers
for CBCT-based detection of TMJ osteoarthritis under a low-data
regime. Across DINO-family self-supervised models (DINOv1, DINOv2,
DINOv2+reg, RAD-DINO) and a supervised ImageNet baseline, fully
frozen representations perform poorly while controlled partial
fine-tuning of the final transformer blocks consistently improves
performance. DINOv2 with partial unfreezing achieves the highest AUC (0.902).
A controlled resolution experiment shows that DINOv2 at 224\,px
matches 518\,px performance after partial adaptation (0.900 vs.\
0.902), confirming that adaptation strategy---not input resolution---
is the dominant performance driver.
Notably, domain-specific RAD-DINO outperforms frozen general-purpose
models without fine-tuning, yet still falls short of partially
adapted DINOv2, underscoring that adaptation strategy is as
important as pretraining domain in low-data medical imaging.

Our experiments further show that attention-based MIL aggregation
and axial slice selection are key design choices, and that 12 slices
suffice---dense volumetric sampling is not necessary.
Future work includes validation on larger datasets, exploration of
hybrid 2D--3D architectures, partial adaptation of RAD-DINO, and
extension toward severity grading and longitudinal monitoring.
These findings highlight that effective adaptation strategies are
critical for leveraging foundation models in low-data medical
imaging settings.


\end{document}